# The utility of tactile force to autonomous learning of in-hand manipulation is task-dependent

Romina Mir, Ali Marjaninejad, Francisco J. Valero-Cuevas*, *Senior Member, IEEE*

*Abstract*—Tactile sensors provide information that can be used to learn and execute manipulation tasks. Different tasks, however, might require different levels of sensory information; which in turn likely affect learning rates and performance. This paper evaluates the role of tactile information on autonomous learning of manipulation with a simulated 3-finger tendon-driven hand. We compare the ability of the same learning algorithm (Proximal Policy Optimization, PPO) to learn two manipulation tasks (rolling a ball about the horizontal axis with and without rotational stiffness) with three levels of tactile sensing: no sensing, 1D normal force, and 3D force vector. Surprisingly, and contrary to recent work on manipulation, adding 1D force-sensing did not always improve learning rates compared to no sensing—likely due to whether or not normal force is relevant to the task. Nonetheless, even though 3D force-sensing increases the dimensionality of the sensory input—which would in general hamper algorithm convergence—it resulted in faster learning rates and better performance. We conclude that, in general, sensory input is useful to learning only when it is relevant to the task—as is the case of 3D force-sensing for in-hand manipulation against gravity. Moreover, the utility of 3D force-sensing can even offset the added computational cost of learning with higher-dimensional sensory input.

## I. INTRODUCTION

Tactile sensing and perception are needed to provide information about contact forces and object properties [1]–[3]. Dexterous manipulation (e.g., in-hand rolling and sliding, finger gaits, re-grasping) has been an active area of research for decades [4]–[6]. Importantly, grasp is not manipulation [6] because the former simply couples an object to the hand, whereas the latter involves the dynamics of changing the orientation of the object with respect to the fingers and fingertips. In-hand manipulation, in which an object is moved while being held, is more challenging and likely depends on the availability of proper tactile information [5], [6]. Moreover, precise object handling requires information about normal and tangential forces, and the slip between the finger and the object [7]–[9]. Even though there is strong evidence that tangential forces play a role in dexterous manipulation tasks in humans [7], [10], [11], the contribution of these tangential forces to autonomous dexterous manipulation in robots remains unexplored. To our knowledge, only one study has explored improved autonomous dexterous in-hand manipulation while the object rested on the palm of the upturned hand [12].

We are particularly interested in understanding the actual control problem the nervous system faces when controlling tendon-driven hands (which are simultaneously under- and over-determined at the same time [13], [14]), and the application of that knowledge to improve bio-inspired robotic systems [15]–[19]. In-hand manipulation is essential for social and personal assistive robots to perform in the human environment [8], [20]. We, therefore, used a tendon-driven hand to help us understand the role of sensory input for the control of in-hand manipulation and future tendon-driven hand prostheses.

To this end, we designed a bio-inspired tendon-driven hand to perform autonomous learning for in-hand manipulation in simulation using different levels of sensory feedback. Not knowing the optimal type and distribution of tactile sensors on the fingers or object, and the need to process large data sets are some of the reasons why sensing has not made systematic headway in robotic manipulation [21]. A bio-inspired approach may inform how to raise the level of tactile sensitivity and versatility in robotic and prosthetic manipulation while performing in human environments [13], [22], [23].

In this paper, we compare the performance of a simulated three-finger hand to achieve two dynamical manipulation tasks: rolling a ball suspended from the fingers against gravity, with (Task-1) and without (Task-2) the need to overcome rotational stiffness along a horizontal rotation axis, Fig. 1. This task was learned in three sensory conditions: with no sensory feedback, with 1D normal fingertip force, and with 3D fingertip forces.

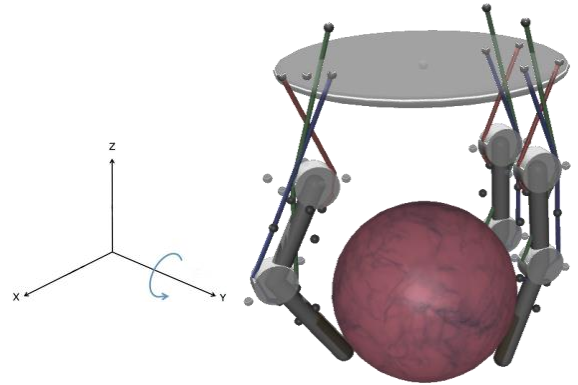

Figure 1. The three-finger tendon-driven hand in the MuJoCo environment, holding a ball against gravity and rotating it about the Y axis.

R. M. is graduated from University of California San Diego, San Diego, Ca, 92093 (e-mail: romirtab@ucsd.edu).
A. M. is with University of Southern California, Los Angeles, Ca, 90089 (e-mail: marjanin@usc.edu).
F. J. V. is with University of Southern California, Los Angeles, Ca, 90089 (corresponding author*; phone: (213) 740-4219; fax: (213) 821-5696; e-mail: valero@usc.edu).

## II. METHODS

Our simulated two-joint, three-tendon fingers in the MuJoCo physics environment [24] (Fig. 1 and see the Supplementary Information section for the video) uses a contractile element with Force-Length-Velocity properties (MuJoCo's built-in muscle actuator). We also use MuJoCo's built-in features to record 1D force ('touch') and 3D force sensing ('Force') on the fingertips of all three fingers [4], [24]. 'Touch' sensor sites provide a nonnegative scalar-value indicating the normal contact force, and 'Force' sensor sites provide a 3D array of 3 orthogonal forces (one normal and two tangential to the sensor site for each sensor) of scalar values representing the 3D contact force vector. The sliding-friction coefficient for dynamically generated contact pairs is also specified for all fingertips (a.k.a. soft contact with friction) [24].

As to the learning algorithm, we used the end-to-end Proximal Policy Optimization (PPO) autonomous learning algorithm [22] as per the PPO1 implementation from OpenAI's stable baselines repository with MultiLayer Perceptron (MLP) Artificial Neural Network (ANN) as for the actor-critic map. We ran the training for 100 episodes (1,000 samples each, for a total of $1*10^5$ samples) for Task-1 and 1,000 episodes for Task-2 (also 1,000 samples each, for a total of $1*10^6$ samples). These numbers correspond to where the learning curves for each task started to flatten (plateau) in preliminary runs.

Hand: The simulated hand consists of a palm and 3 identical tendon-driven fingers: two adjacent fingers (analogous to the 'index' and 'middle' fingers) and one opposing them (analogous to the 'thumb') (Fig. 1). Each finger consists of two joints. The size of the palm and length-ratio of each 'phalanx' was based on the human hand ratios tactile [12]. The ranges of flexion-extension of the joints are loosely based on those of the human hand [12], [25], [26]. The fingers do not have an abduction-adduction degree of freedom at their base and can only flex or extend at the two joints. The sensory sites are only used on the internal side (i.e., the 'pads of the fingertips') of the distal phalanx of each finger.

Ball: We use a ball with friction as our experimental object in both Tasks (see Fig. 1). The ball has a mass of 2.8 kg and it has a total of 4-degrees of freedom (DoF): three translational DoFs (X, Y, and Z) with a built-in spring-damper in all three directions-a time constant of 0.09 seconds and being critically damped (damping coefficient=1) and one rotational degree of freedom along the Y-axis.

Both Tasks are defined as holding the ball against gravity and rotating it about the Y-axis in the positive direction (shown in Fig. 1). In Task-1 there is no friction or stiffness to resist this rotation, whereas in Task-2 the rotation is frictionless but happens against a torsional spring with the stiffness resistance of value 30K (in arbitrary units of MuJoCo, and found heuristically by us) [24]. In both Tasks, the goal is to rotate the ball as far as possible in a 10-second trial. Since this task requires the ball to be rotated against a torsional spring that tends to rotate the ball back, any successful solution will involve both rotating the ball while maintaining force closure to prevent it from returning to its original position. As mentioned above, each Task was performed in three sensory conditions:

No Force: Using No sensory information

1D Force: Using 1D Normal Force information.

3D Force: Using 3D Force vector information.

## III. RESULTS

Figure 2 shows reward vs. episode plots for 100 Monte Carlo runs of each task. The reward itself is task dependent, but the learning curves for both tasks exhibit a consistent pattern of initial rapid improvement followed by a more gradual, less steep improvement. Figure 2 shows the 3D Force sensory condition outperforms the others in both Tasks, as could be expected. More interestingly, 1D Force performs only marginally better than No Force for Task-1 yet performs worse than both other conditions for Task-2. These results suggest that for autonomous in-hand manipulation (i) more data are not always better and, (ii) if relevant, more and higher-dimensional sensory data do not necessarily slow down learning even though more free parameters (weights) need to be adjusted on the ANN used for the actor-critic part of the PPO algorithm. Note the difference in the X-axis of the plots in the figures below.

Figure 3 shows boxplots and means of the final reward for 100 Monte Carlo runs for both experiments with total training of $1*10^5$ samples for Task-1 and $1*10^6$ samples for Task-2 for each run. The final reward for Task-1 shows a gradual increase progressing from No to 1D to 3D Force sensing conditions. However, increasing the sensory information does not show this pattern for Task-2: there is a decrease of the two No to 1D Force sensing conditions. Pairwise two-sided Student t-tests (which may be of limited accuracy due to lack

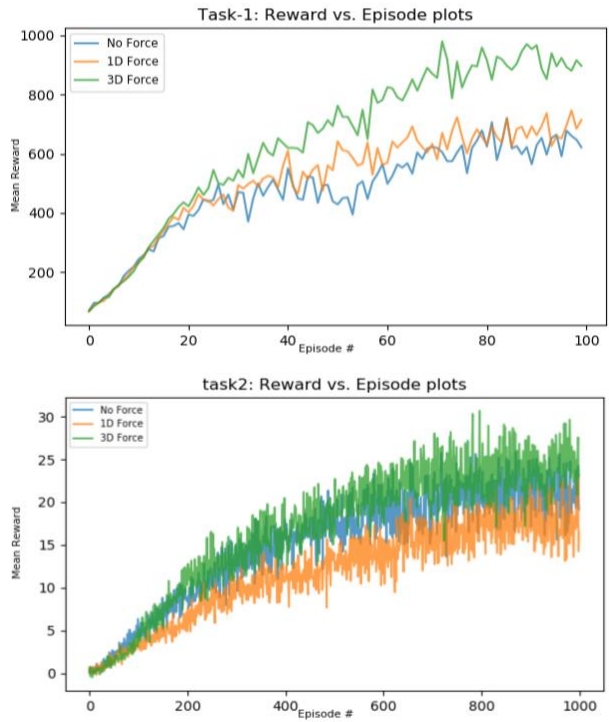

Figure 2. Results of the manipulation tasks using the PPO algorithm, 100 Monte Carlo runs for each case.

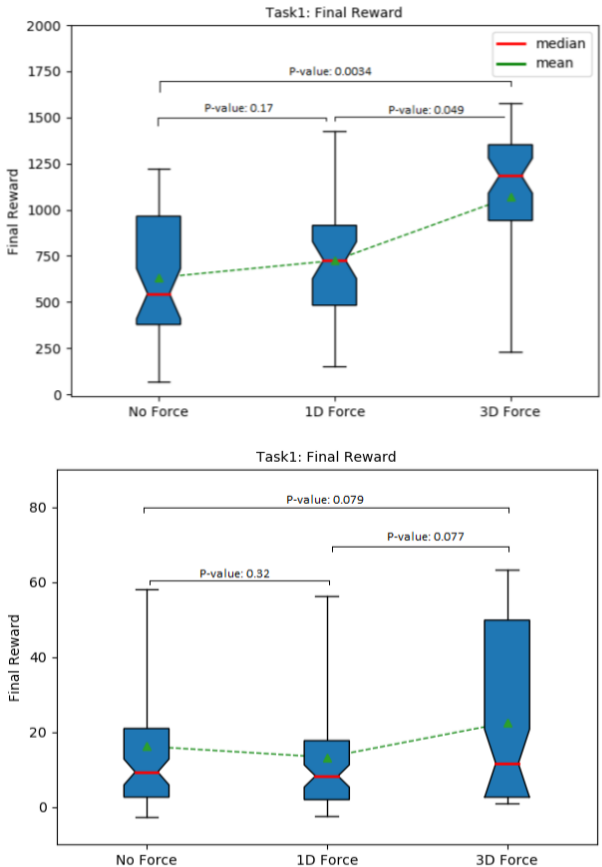

Fig. 3: Boxplot of Final rewards with the mean, using PPO algorithm with 100 Monte Carlo runs for each case (1*10^5 episode for Task-1 and 1*10^6 episode for Task-2 for each Monte Carlo run). P-values are also of homoscedasticity) among these independently run simulations with the three sensory conditions shows that the terminal reward of the No and 1D Force conditions for both Tasks do not show significant differences ($p < 0.17$ in Task-1 and $p < 0.32$ in Task-2). In contrast, the terminal costs of the 3D Force condition were better than No Force and 1D Force for Task-1 ($p < 0.0034$ and $p < 0.049$, respectively), and close to significantly better for Task-2 ($p < 0.079$ and $p < 0.077$, respectively).central quartiles and mean reward from

## IV. Discussion

In most in-hand manipulation tasks with soft fingers [16], the force between the fingertip and the object is not oriented normal to the surface of the sensor assembly. In biological skin, mechanoreceptors sense the stress and strain distributions within the fingertip induced by contact [20]. We find that the inclusion of 3D Force information improves model-free deep reinforcement learning, despite the increased dimensionality of the dataset and resulting added free parameters in the learning algorithm. These results underline the importance and utility of 3D endpoint force sensing in robotic hands for in-hand dexterous manipulation tasks.

Tactile sensing is a crucial source of information for manipulation of objects and tool usage for humans and robots [12], [27], [28]. While recent work [12] shows the performance of the agent can be increased when normal 1D Force information is available to the agent, we find this is not always the case. One of the main differences between our Tasks and the one studied in [12] (where the object was manipulated on the upturned palm) is that Tasks-1 and -2 both required force closure to overcome gravity, in addition to force closure to counteract the rotational spring in Task-2. For example, we sometimes see the ball drops from the hand while learning/trying to rotate the ball. Our results suggest that feedback using information that is irrelevant to a task (i.e., 1D Force) can be as much of an impediment (or have no added value) to learning and performance as the lack of information (i.e., No Force). That is, 1D Force is not necessarily informative of impending slip due to gravity or the torsional spring and produced no improvement over No Force.

Figure 3 shows a high variability in the final reward for both Tasks. Two possible sources are: (i) Starting the RL algorithm from random initial states tends to find alternative local minima which can be dispersed if the fitness landscape is shallow, which is common in some RL problems; and (ii) both Tasks deal with high-level dynamics including intermittent contact, linear and rotational inertia, and multiple degrees of freedom—all of which allows for a wide spectrum of solutions. Therefore, the high variability in the results is not surprising (see the supplementary video).

In future work, we will expand our study of the effects of sensory levels on a wide spectrum of tasks across the taxonomy of in-hand manipulation [29]. This will help clarify the co-evolution of sensory and motor systems for biological manipulation, and the utility of sensory information to co-optimize the type of sensors built into robotics hands with the learning algorithms for autonomous dexterous manipulation.

## Supplementary Information

The code and the supplementary files (Video, MuJoCo models, etc.) which is used in this study can be accessed through its GitHub repository at: https://github.com/rominamir/sensorylevel

## Acknowledgment

Research reported in this publication was supported in part by the National Institute of Arthritis and Musculoskeletal and Skin Diseases of the National Institutes of Health under award number R01 AR-050520 and R01 AR-052345 and by the Department of Defense CDMRP Grant MR150091 and Award W911NF1820264 from the DARPA-L2M program to F.J.V.-C. This work does not necessarily represent the views of the NSF, NIH, DoD, or DARPA.